\documentclass{article}
\usepackage{iclr2020_conference,times}

\usepackage{amsmath,amsfonts,bm}

\def\eqref#1{equation~\ref{#1}}

\def\1{\bm{1}}

\DeclareMathAlphabet{\mathsfit}{\encodingdefault}{\sfdefault}{m}{sl}
\SetMathAlphabet{\mathsfit}{bold}{\encodingdefault}{\sfdefault}{bx}{n}

\usepackage{hyperref,url}
\usepackage{amsmath}
\usepackage{booktabs}
\usepackage{amsfonts}
\usepackage{nicefrac}
\usepackage{microtype}

\usepackage{graphicx}
\usepackage{algorithmicx,algorithm}
\usepackage{wrapfig}
\usepackage{amsmath}
\usepackage{tikz}
\usetikzlibrary{arrows}
\usetikzlibrary{calc,patterns}
\usepackage{pdflscape}
\title{DyNet: Dynamic Convolution for Accelerating Convolutional Neural Networks}
\iclrfinalcopy

\author{Yikang Zhang, Jian Zhang, Qiang Wang, Zhao Zhong \\
	HUAWEI\\
	\texttt{\{zhangyikang7, zhangjian157, zorro.zhongzhao\}@huawei.com} \\
}

\begin{document}
\maketitle

\begin{abstract}
Convolution operator is the core of convolutional neural networks (CNNs) and occupies the most computation cost. To make CNNs more efficient, many methods have been proposed to either design lightweight networks or compress models. Although some efficient network structures have been proposed, such as MobileNet or ShuffleNet, we find that there still exists redundant information between convolution kernels. To address this issue, we propose a novel \textbf{dynamic convolution} method to adaptively generate convolution kernels based on image contents. To demonstrate the effectiveness, we apply dynamic convolution on multiple state-of-the-art CNNs. On one hand, we can reduce the computation cost remarkably while maintaining the performance. For ShuffleNetV2/MobileNetV2/ResNet18 /ResNet50, DyNet can reduce $37.0/54.7/67.2/71.3\%$ FLOPs without loss of accuracy. On the other hand, the performance can be largely boosted if the computation cost is maintained. Based on the architecture MobileNetV3-Small/Large, DyNet achieves $70.3/77.1\%$ Top-1 accuracy on ImageNet with an improvement of $2.9/1.9\%$. To verify the scalability, we also apply DyNet on segmentation task, the results show that DyNet can reduce $69.3\%$ FLOPs while maintaining Mean IoU on segmentation task.
\end{abstract}

\section{Introduction}\label{intro}
Convolutional neural networks (CNNs) have achieved state-of-the-art performance in many computer vision tasks \citep{krizhevsky2012imagenet,szegedy2013deep}, and the neural architectures of CNNs are evolving over the years \citep{krizhevsky2012imagenet,simonyan2014very,szegedy2015going,he2016deep,hu2018squeeze,zhong2018practical,zhong2018blockqnn}. However, modern high-performance CNNs often require a lot of computation resources to execute a large amount of convolution kernel operations. Aside from the accuracy, to make CNNs applicable on mobile devices, building lightweight and efficient deep models has attracting much more attention recently \citep{howard2017mobilenets,sandler2018mobilenetv2,howard2019searching,zhang2018shufflenet,ma2018shufflenet}. These methods can be roughly categorized into two types: efficient network design and model compression. Representative methods for the former category are MobileNet \citep{howard2017mobilenets,sandler2018mobilenetv2,howard2019searching} and ShuffleNet \citep{ma2018shufflenet,zhang2018shufflenet}, which use depth-wise separable convolution and channel-level shuffle techniques to reduce computation cost. On the other hand, model compression-based methods tend to obtain a smaller network by compressing a larger network via pruning, factorization, mimic and quantization \citep{chen2015compressing, han2015deep, jaderberg2014speeding, lebedev2014speeding,ba2014deep,zhu2016trained}.

Although some handcrafted efficient network structures have been designed, we observe that the significant correlations still exist among convolutional kernels, and introduce a large amount of redundant calculations. Moreover, these small networks are hard to compress. For example, \citet{liu2019metapruning} compress MobileNetV2 to 124M, but the accuracy drops by $5.4\%$ on ImageNet compared with MobileNetV2 (1.0). This implies that traditional compression methods cannot solve the inherent redundancy problem in CNNs well. We theoretically analyze this phenomenon and find that it is caused by the nature of conventional convolution, where correlated kernels are cooperated to extract noise-irrelevant features. Thus it is hard to compress the conventional convolution kernels without information loss. We also find that if we linearly fuse several fixed convolution kernels to generate one dynamic kernel based on the input, we can obtain the noise-irrelevant features without the cooperation of multiple kernels.

\begin{figure}[t]
	\centering
	\includegraphics[width=0.9\linewidth]{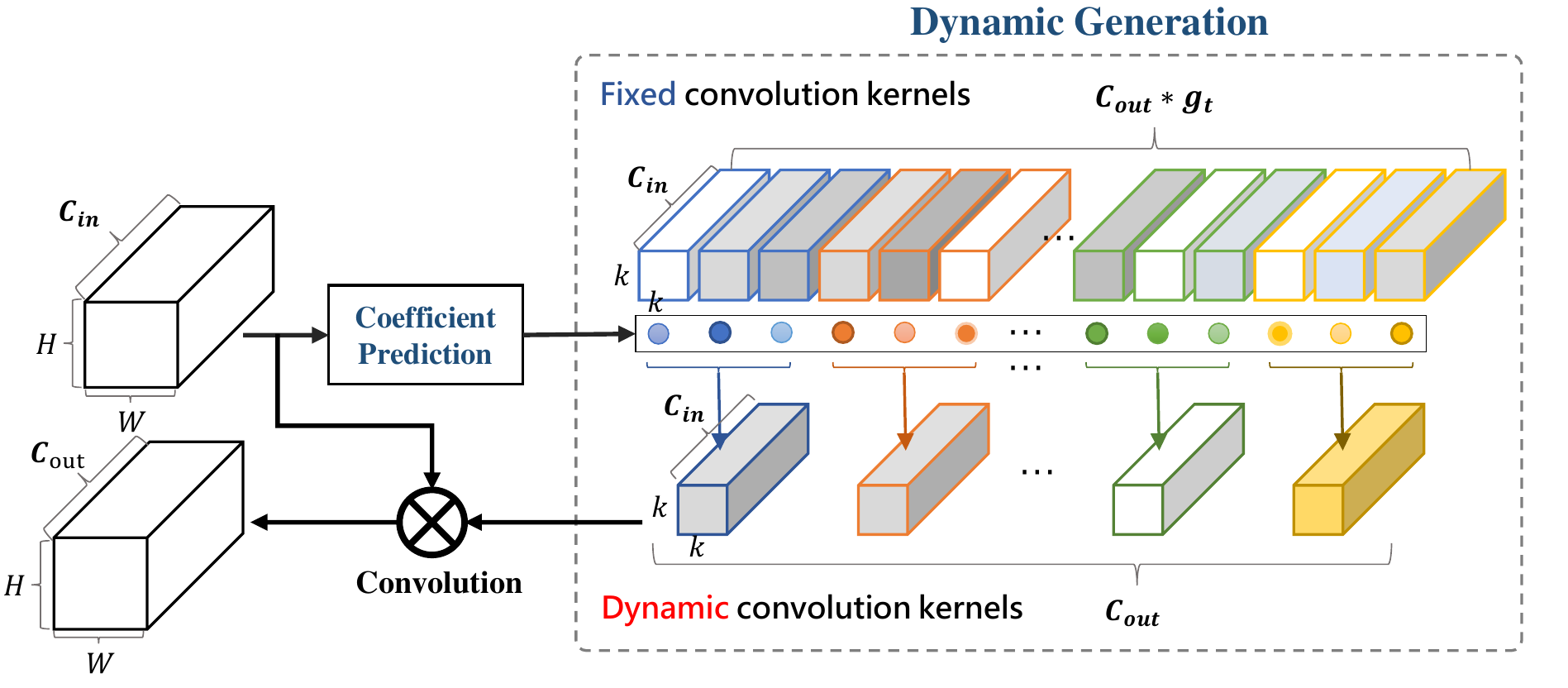}
	\caption{The overall framework of the dynamic convolution.}
	\label{dynamic_conv_layer}
\end{figure}

Based on the above observation and analysis, we propose dynamic convolution to address this issue, which adaptively generates convolution kernels based on image contents. The overall framework of dynamic convolution is shown in Figure \ref{dynamic_conv_layer}, which consists of a \textit{coefficient prediction module} and a \textit{dynamic generation module}. The coefficient prediction module is trainable and designed to predict the coefficients of fixed convolution kernels. Then the dynamic generation module further generates a dynamic kernel based on the predicted coefficients.

Our proposed method is simple to implement and can be used as a drop-in plugin for any convolution layer to reduce redundancy. We evaluate the proposed dynamic convolution on state-of-the-art networks. On one hand, we can reduce the computation cost remarkably while maintaining the performance. For ShuffleNetV2 (1.0), MobileNetV2 (1.0), ResNet18 and ResNet50, DyNet reduces $37.0\%$, $54.7\%$, $67.2\%$ and $71.3\%$ FLOPs respectively while the Top-1 accuracy on ImageNet changes by $+1.0\%$, $-0.27\%$, $-0.6\%$ and $-0.08\%$. On the other hand, the performance can be largely boosted if the computation cost is maintained. For MobileNetV3-Small(1.0) and MobileNetV3-Large(1.0), DyNet improve the Top-1 accuracy on ImageNet by $2.9\%$ and $1.9\%$ respectively while the FLOPs changes by
$+4.1\%$ and $+5.3\%$. Meanwhile, dynamic convolution further accelerates the inference speed of MobileNetV2 (1.0), ResNet18 and ResNet50 by 1.87$\times$,1.32$\times$and 1.48$\times$ on CPU platform respectively.

\section{Related Work}\label{related_work}
We review related works from three aspects: efficient convolution neural network design, model compression, and dynamic convolution kernel.

\subsection{Efficient convolution neural network design}
In many computer vision tasks \citep{krizhevsky2012imagenet,szegedy2013deep}, model design plays a key role. The increasing demands of high-quality networks on mobile/embedding devices have driven the study on efficient network design \citep{he2015convolutional}. For example, GoogleNet \citep{szegedy2015going} increases the depth of networks with lower complexity compared to simply stacking convolution layers; SqueezeNet \citep{iandola2016squeezenet} deploys a bottleneck approach to design a very small network; Xception \citep{chollet2017xception}, MobileNet \citep{howard2017mobilenets, sandler2018mobilenetv2} use depth-wise separable convolution to reduce computation and model size. ShuffleNet \citep{zhang2018shufflenet, ma2018shufflenet} shuffle channels to reduce the computation of $1\times1$ convolution kernel and improve accuracy. MobileNetV3 \cite{howard2019searching} are designed based on a combination of complementary search techniques. Despite the progress made by these efforts, we find that there still exists redundancy between convolution kernels and cause redundant computation. Dynamic convolution can reduce the redundant computation, thus complement those efficient networks.

\subsection{Model compression}
Another trend to obtaining a small network is model compression. Factorization based methods \citep{jaderberg2014speeding,lebedev2014speeding} try to speed up convolution operation by using tensor decomposition to approximate original convolution operation.  Knowledge distillation based methods \citep{ba2014deep,romero2014fitnets,hinton2015distilling} learn a small network to mimic a larger teacher network. Pruning based methods \citep{han2015deep,han2015learning,wen2016learning,liu2019metapruning} try to reduce computation by pruning the redundant connections or convolution channels. Compared with those methods, DyNet is more effective especially when the target network is already efficient enough. For example, in \citep{liu2019metapruning}, they get a smaller model of 124M FLOPs by pruning the MobileNetV2, however, it drops the accuracy by $5.4\%$ on ImageNet compared with the model with 291M FLOPs. Moreover, the pruned MobileNetV2 with 137M FLOPs in \citep{ye2020good} drops the accuracy by $3.2\%$ and the pruned ResNet50 with 2120M FLOPs in \citep{wang2019pruning} drops the accuracy by $5.1\%$. While in DyNet, we can reduce the FLOPs of MobileNetV2 (1.0) from 298M to 129M with the accuracy drops only $0.27\%$ and reduce the FLOPs of ResNet50 from 3980M to 1119M with the accuracy drops only $0.08\%$.

\subsection{Dynamic convolution kernel}
Generating dynamic convolution kernel appears in both computer vision and natural language processing (NLP) tasks. 

In computer vision domain, Klein et al. \citep{klein2015dynamic} and Brabandere et al. \citep{jia2016dynamic} directly generate convolution kernels via a linear layer based on the feature maps of previous layers. Because convolution kernels have a large number of parameters, the linear layer will be inefficient on the hardware. Our proposed method solves this problem by merely predicting the coefficients for linearly combining fixed kernels and achieve real speed up for CNN on hardware. This technique has been deployed in HUAWEI at the beginning of 2019 and the patent is filed in May 2019 as well. The attention paid by the academic community \citep{yang2019soft, chen2019dynamic, chen2020dynamic} demonstrates the great potential of this direction. In this paper, we derive insight into dynamic convolution from the perspective of 'noise-irrelevant feature' and conduct a correlation experiment to prove that the correlation among convolutional kernels can be largely reduced in DyNet.

In NLP domain, some works \citep{shen-etal-2018-learning,wu2019pay,gong2018convolutional} incorporate context information to generate input-aware convolution filters which can be changed according to input sentences with various lengths. These methods also directly generate convolution kernels via a linear layer, etc. Because the size of CNN in NLP is smaller and the dimension of the convolution kernel is one, the inefficiency issue for the linear layer is alleviated.
Moreover, Wu et al. \citep{wu2019pay} alleviate this issue utilizing the depthwise convolution and the strategy of sharing weight across layers. These methods are designed to improve the adaptivity and flexibility of language modeling, while our method aims to cut down the redundant computation cost.

\section{DyNet: Dynamic Convolution in CNNs}\label{dck}
In this section, we first describe the motivation of dynamic convolution. Then we explain the proposed dynamic convolution in detail. Finally, we illustrate the architectures of our proposed DyNet.

\begin{figure}[htb]
	\centering
	\includegraphics[width=0.6\linewidth]{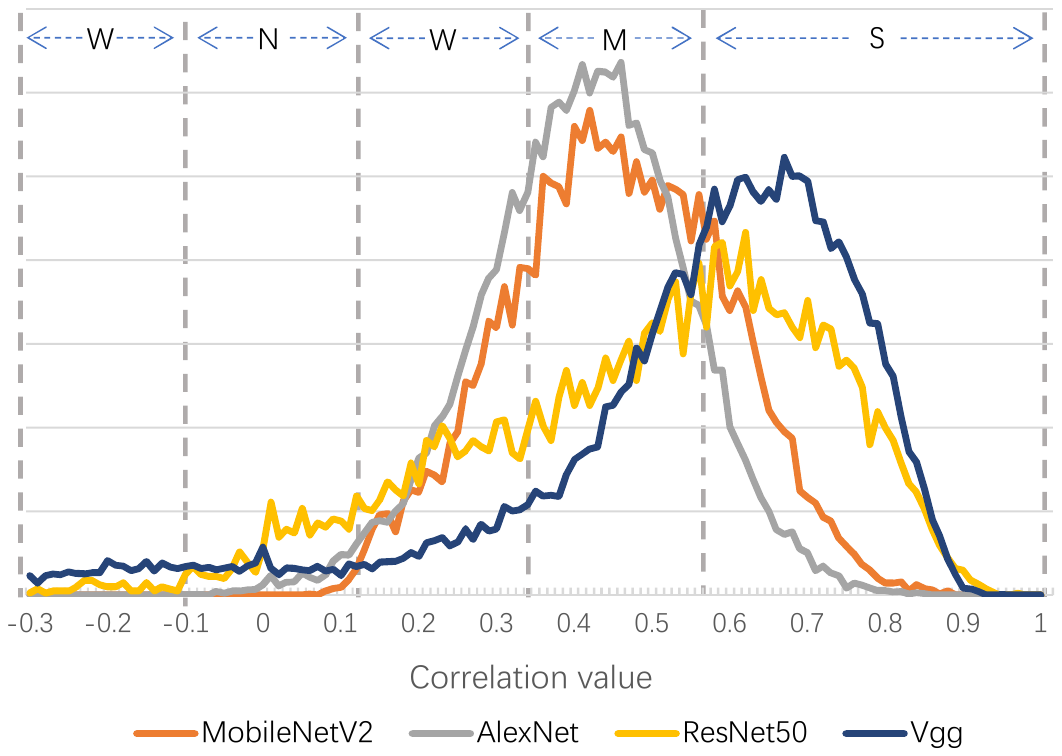}
	\caption{Pearson product-moment correlation coefficient between feature maps. S, M, W, N denote strong, middle, weak and no correlation respectively.}
	\label{Correlation}
\end{figure}

\subsection{Motivation}
As indicated in previous works~\citep{han2015deep,han2015learning,wen2016learning,liu2019metapruning}, convolutional kernels are naturally correlated in deep models. For some of the well-known networks, we plot the distribution of the Pearson product-moment correlation coefficient between feature maps in Figure~\ref{Correlation}. Most existing works try to reduce correlations by compressing, however, it is hard to accomplish for efficient and small networks like MobileNets, even though the correlation is significant. We think these correlations are vital for maintaining the performance because they are cooperated to obtain noise-irrelevant features. We take face recognition as an example, where the pose or the illumination is not supposed to change the classification results. Therefore, the feature maps will gradually become noise-irrelevant when they go deeper. Based on the theoretical analysis in appendix~\ref{Appendix}, we find this procedure needs the cooperation of multiple correlated kernels and we can get noise-irrelevant features without this cooperation if we dynamically fuse several kernels. In this paper, we propose a dynamic convolution method, which learns the coefficients to fuse multiple kernels into a dynamic one based on image contents. We give a more in-depth analysis of our motivation in appendix~\ref{Appendix}.

\subsection{Dynamic convolution}
The goal of dynamic convolution is to learn a group of kernel coefficients, which fuse multiple fixed kernels to a dynamic one. We illustrate the overall framework of dynamic convolution in Figure \ref{dynamic_conv_layer}. We first utilize a trainable coefficient prediction module to predict coefficients. Then we further propose a dynamic generation module to fuse fixed kernels to a dynamic one. We will introduce the coefficient prediction module and dynamic generation module in detail in the following of this section.

\paragraph{Coefficient prediction module}
\begin{wrapfigure}{R}{0.5\linewidth}
\centering
\includegraphics[width=\linewidth]{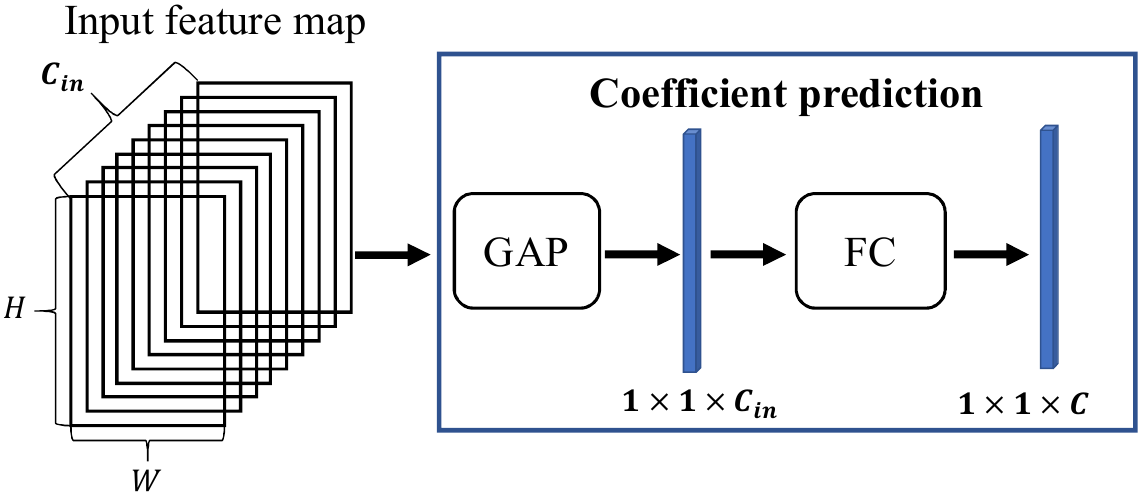}
\caption{The coefficient prediction module.}
\label{redundancy_predict}
\end{wrapfigure}

Coefficient prediction module is proposed to predict coefficients based on image contents. As shown in Figure~\ref{redundancy_predict}, the coefficient prediction module can be composed of a global average pooling layer and a fully connected layer with Sigmoid as activation function. Global average pooling layer aggregates the input feature maps into a $1\times 1\times C_{in}$ vector, which serves as a feature extraction layer. Then the fully connected layer further maps the feature into a $1 \times 1\times C$ vector, which are the coefficients for fixed convolution kernels of several dynamic convolution layers.

\paragraph{Dynamic generation module}
For a dynamic convolution layer with weight $[C_{out}\times g_t,C_{in},k,k]$, it corresponds with $C_{out} \times g_t$ fixed kernels and $C_{out}$ dynamic kernels, the shape of each kernel is $[C_{in},k,k]$. $g_t$ denotes the group size, it is a hyperparameter. We denote the fixed kernels as $w_t^i$, the dynamic kernels as $\widetilde{w_t}$, the coefficients as $\eta_t^i$, where $t=0,...,C_{out},i=0,...,g_t$.

After the coefficients are obtained, we generate dynamic kernels as follows:
\begin{equation}
\label{dy_conv}
\widetilde{w}_t = \sum_{i=1}^{g_t}\eta_{t}^i\cdot w_t^i
\end{equation}

\paragraph{Training algorithm} For the training of the proposed dynamic convolution, it is not suitable to use the batch-based training scheme. It is because the convolution kernel is different for different input images in the same mini-batch. Therefore, we fuse feature maps based on the coefficients rather than kernels during training. They are mathematically equivalent as shown in Eq. \ref{eq9}:
\begin{equation}
\begin{split}
\widetilde{O}_t &= \widetilde{w}_t \otimes x = \sum_{i=1}^{g_t}(\eta_{t}^i\cdot w_t^i) \otimes x = \sum_{i=1}^{g_t}(\eta_{t}^i\cdot w_t^i \otimes x )\\
&= \sum_{i=1}^{g_t}(\eta_{t}^i\cdot (w_t^i \otimes x))= \sum_{i=1}^{g_t}(\eta_{t}^i\cdot {O}_t^i),\label{eq9}
\end{split}
\end{equation}
where $x$ denotes the input, $\widetilde{O}_t$ denotes the output of dynamic kernel $\widetilde{w}_t$, ${O}_t^i$ denotes the output of fixed kernel ${w}_t^i$.

\begin{figure}[htb]
\centering
\includegraphics[width=0.9\linewidth]{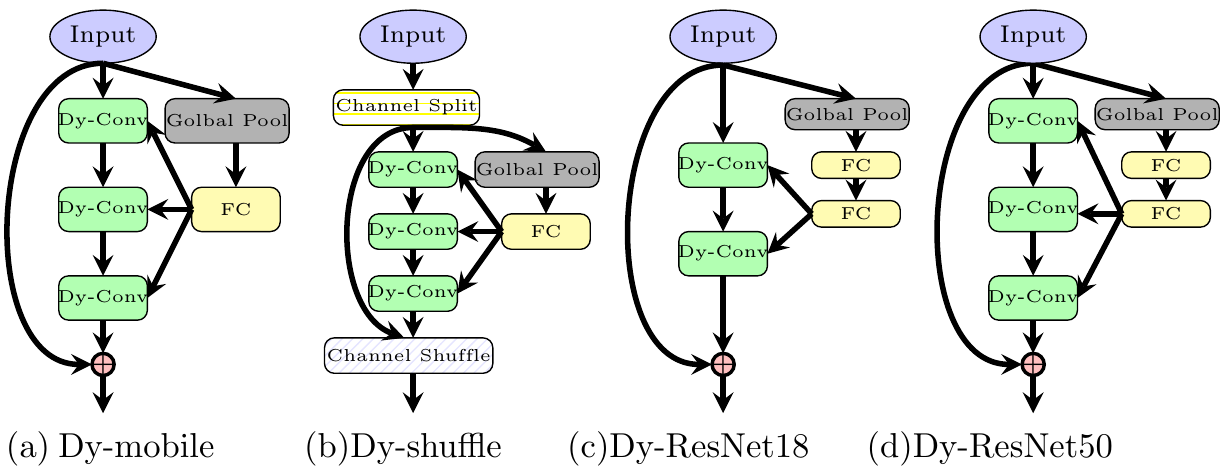}
\caption{Basic building bolcks for Dynamic Network variants of MobileNet, ShuffleNet, ResNet18, and ResNet50.}
\label{Dy___Nets}
\end{figure}

\subsection{Dynamic convolution neural networks}
We equip MobileNetV2, ShuffleNetV2, and ResNets with our proposed dynamic convolution, and propose Dy-mobile, Dy-shuffle, Dy-ResNet18, and Dy-ResNet50 respectively. The building blocks of these 4 networks are shown in Figure~\ref{Dy___Nets}. Based on dynamic convolution, each dynamic kernel can get a noise-irrelevant feature without the cooperation of other kernels. Therefore we can reduce the channels for DyNets and remain the performance. We set the hyper-parameter $g_t$ as 6 for all of them, and we give details of these dynamic CNNs below.
To verify the performance can also be largely boosted if the number of channels is kept, we simply replace the convolution of MobileNetV3-Small(1.0) and MobileNetV3-Large(1.0) with the dynamic one to get the Dy-MobileNetV3-Small and Dy-MobileNetV3-Large.

\paragraph{Dy-mobile}
In our proposed Dy-mobile, we replace the original MobileNetV2 block with our dy-mobile block, which is shown in Figure~\ref{Dy___Nets} (a). The input of coefficient prediction module is the input of block, it produces the coefficients for all three dynamic convolution layers. Moreover, we further make two adjustments:

\begin{itemize}
\item We do not expand the channels in the middle layer like MobileNetV2. If we denote the output channels of the block as $C_{out}$, then the channels of all the three convolution layers will be $C_{out}$.
\item Since the depth-wise convolution is efficient, we set $\mathit{groups}=\frac{C_{\mathit{out}}}{6}$ for the dynamic depth-wise convolution. We will enlarge $C_{\mathit{out}}$ to make it becomes the multiple of 6 if needed.
\end{itemize}

After the aforementioned adjustments, the first dynamic convolution layer reduces the FLOPs from $6C^2HW$ to $C^2HW$. The second dynamic convolution layer keeps the FLOPs as $6CHW \times 3^2$ unchanged because we reduce the output channels by 6x while setting the groups of convolution 6x smaller, too. For the third dynamic convolution layer, we reduce the FLOPs from $6C^2HW$ to $C^2HW$ as well. The ratio of FLOPs for the original block and our dy-mobile block is:
\begin{equation}
\begin{split}
\frac{6C^2HW+6CHW\times 3^2+6C^2HW}{C^2HW+6CHW\times 3^2+C^2HW}=\frac{6C+27}{C+27}=6-\frac{135}{C+27}
\end{split}
\end{equation}

\paragraph{Dy-shuffle}
In the original ShuffleNet V2, channel split operation will split feature maps to right-branch and left-branch, the right branch will go through one pointwise convolution, one depthwise convolution, and one pointwise convolution sequentially. We replace conventional convolution with dynamic convolution in the right branch as shown in Figure~\ref{Dy___Nets} (b). We feed the input of the right branch into coefficient prediction module to produce the coefficients. In our dy-shuffle block, we split channels into left-branch and right-branch with ratio $3:1$, thus we reduce the $75\%$ computation cost for two dynamic pointwise convolution. Similar to dy-mobile, we adjust the parameter "groups" in dynamic depthwise convolution to keep the FLOPs unchanged.

\paragraph{Dy-ResNet18/50}
In Dy-ResNet18 and DyResNet50, we simply reduce half of the output channels for dynamic convolution layers of each residual block. Because the input channels of each block are large compared with dy-mobile and dy-shuffle, we use two linear layers as shown in Figure~\ref{Dy___Nets} (c) and Figure~\ref{Dy___Nets} (d) to reduce the number of parameters. If the input channel is $C_{in}$, the output channels of the first linear layer will be $\frac{C_{in}}{4}$ for Dy-ResNet18/50.

\section{Experiments}\label{exp}
\subsection{Implementation details}

For the training of the proposed dynamic neural networks. Each image has data augmentation of randomly cropping and flipping, and is optimized with SGD strategy with cosine learning rate decay. We set batch size, initial learning rate, weight decay and momentum as 2048, 0.8, 5e-5 and 0.9 respectively. We also use the label smoothing with a rate of 0.1. We evaluate the accuracy on the test images with center crop. 

\begin{wrapfigure}{R}{0.5\linewidth}
	\centering
	\setlength{\abovecaptionskip}{-15pt}
	\includegraphics[width=\linewidth]{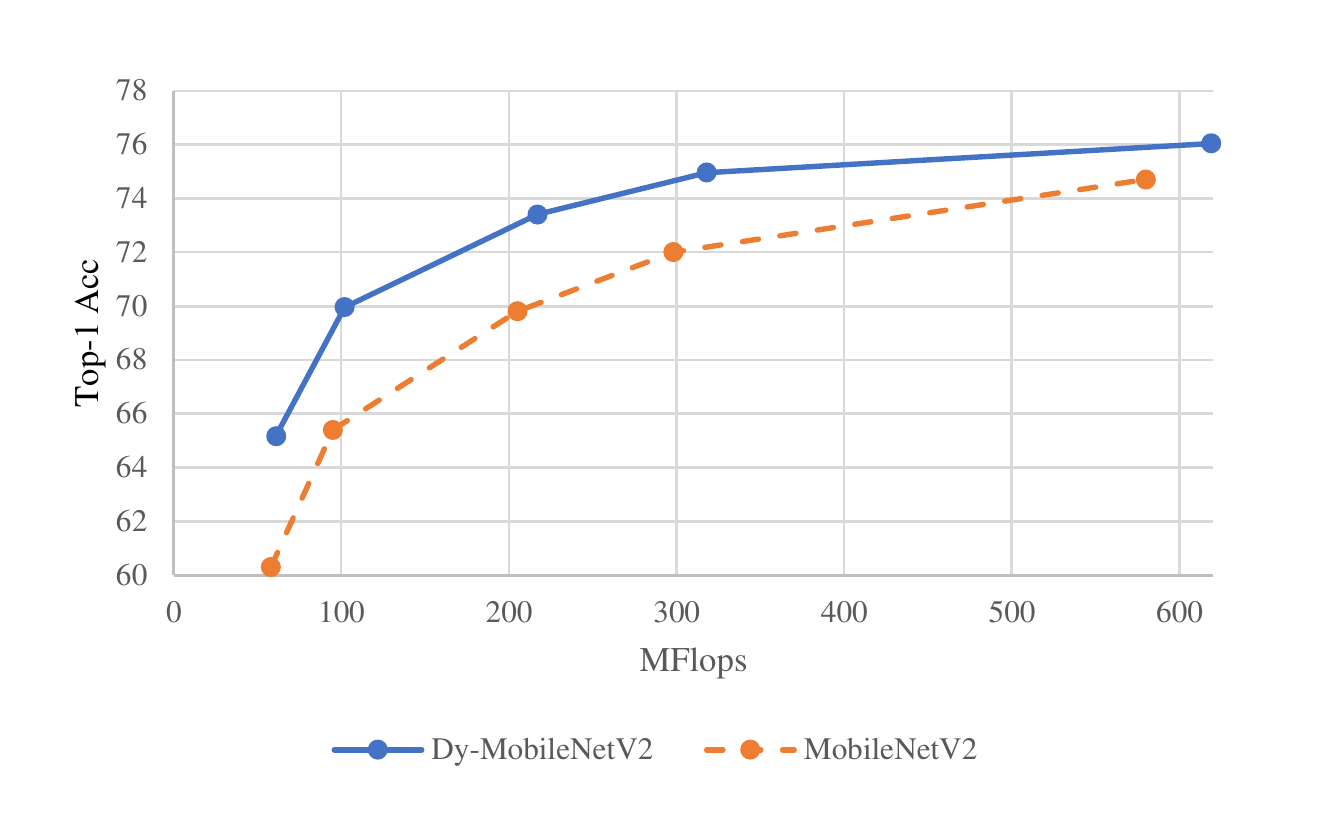}
	\caption{Compare with MobileNetV2 under the similar Flops constraint.}
	\label{mobilenet_results_detail}
\end{wrapfigure}

\subsection{Experiment settings and compared methods}
We evaluate DyNet on ImageNet \citep{russakovsky2015imagenet}, which contains 1.28 million training images and 50K validation images collected from 1000 different classes. We train the proposed networks on the training set and report the top-1 error on the validation set. To demonstrate the effectiveness, we compare the proposed dynamic convolution with state-of-the-art networks under mobile setting, including MobileNetV1~\citep{howard2017mobilenets}, MobileNetV2~\citep{sandler2018mobilenetv2}, ShuffleNet~\citep{zhang2018shufflenet}, ShuffleNet V2~\citep{ma2018shufflenet}, Xception~\citep{chollet2017xception}, DenseNet~\citep{huang2017densely}, IGCV2~\citep{xie2018interleaved} and IGCV3~\citep{sun2018igcv3}.

\begin{table}[tbh]
\caption{Comparison of different network architectures over classification error and computation cost. The number in the brackets denotes the channel number controller \citep{sandler2018mobilenetv2}.}
\label{result_table}
\centering
\begin{tabular}{lll}
\toprule
Methods & MFLOPs & Top-1 err. (\%) \\ 
\midrule
MobileNetV3-Small(1.0)) \citep{howard2019searching} & 56 & 32.60 \\
ShuffleNet V2 (1.0) \citep{ma2018shufflenet} & 146 & 30.60 \\
MobileNetV2 (1.0) \citep{sandler2018mobilenetv2} & 298 & 28.00 \\
MobileNetV3-Large(1.0) \citep{howard2019searching} & 219 & 24.8 \\
ResNet18 & 1730 & 30.41 \\
ResNet50 & 3890 & 23.67 \\
\midrule
ShuffleNet v1 (1.0) \citep{zhang2018shufflenet} & 140 & 32.60 \\
MobileNet v2 (0.75) \citep{sandler2018mobilenetv2} & 145 & 32.10 \\
MobileNet v2 (0.6) \citep{sandler2018mobilenetv2} & 141 & 33.30 \\
MobileNet v1 (0.5)\citep{howard2017mobilenets} & 149 & 36.30 \\
DenseNet (1.0) \citep{huang2017densely} & 142 & 45.20 \\
Xception (1.0) \citep{chollet2017xception} & 145 & 34.10 \\
IGCV2 (0.5) \citep{xie2018interleaved} & 156 & 34.50 \\
IGCV3-D (0.7) \citep{sun2018igcv3} & 210 & 31.50 \\
\midrule
Dy-MobileNetV3-Small & 59 & 29.7 \\
Dy-shuffle (1.0) & 92 & 29.6 \\
Dy-mobile (1.0) & 135 & 28.27 \\
Dy-MobileNetV3-Large & 228 & 22.9 \\
Dy-ResNet18 & 567 & 31.01 \\
Dy-ResNet50 & 1119 & 23.75 \\
\bottomrule
\end{tabular}
\end{table}

\subsection{Experiment results and analysis}

\paragraph{Analysis of accuracy and computation cost} We demonstrate the results in Table \ref{result_table}, where the number in the brackets indicates the channel number controller \citep{sandler2018mobilenetv2}. We partitioned the result table into three parts: (1) The original networks corresponding to the implemented dynamic networks; (2) Compared state-of-the-art networks under mobile settings; (3) The proposed dynamic networks.

Table \ref{result_table} provides several valuable observations: (1) Compared with these well-known models under mobile setting, the proposed Dy-mobile, Dy-shuffle, and Dy-MobileNetV3 achieves the best classification error with lowest computation cost. This demonstrates that the proposed dynamic convolution is a simple yet effective way to reduce computation cost. (2) Compared with the corresponding basic neural structures, the proposed Dy-shuffle (1.0), Dy-mobile (1.0), Dy-ResNet18 and Dy-ResNet50 reduce $37.0\%$, $54.7\%$, $67.2\%$ and $71.3\%$ computation cost respectively with little drop on Top-1 accuracy. 
This shows that even though the proposed network significantly reduces the convolution computation cost, the generated dynamic kernel can still capture sufficient information from image contents. (3) Compared with MobileNetV3-Small(1.0) and MobileNetV3-Large(1.0), Dy-MobileNetV3-Small and Dy-MobileNetV3-Large improve the Top-1 accuracy on ImageNet by $2.9\%$ and $1.9\%$ respectively with the FLOPs only increasing by 3M and 9M. The results also indicate that the performance can be largely boosted if the computation cost is maintained

Furthermore, we conduct detailed experiments on MobileNetV2. We replace the conventional convolution with the proposed dynamic one and get Dy-MobileNetV2. The accuracy of classification for models with different numbers of channels is shown in Figure \ref{mobilenet_results_detail}. It is observed that Dy-MobileNetV2 consistently outperforms MobileNetV2 but the ascendancy is weakened with the increase of the number of channels.

\begin{figure}[htb]
\centering
\includegraphics[width=0.8\linewidth]{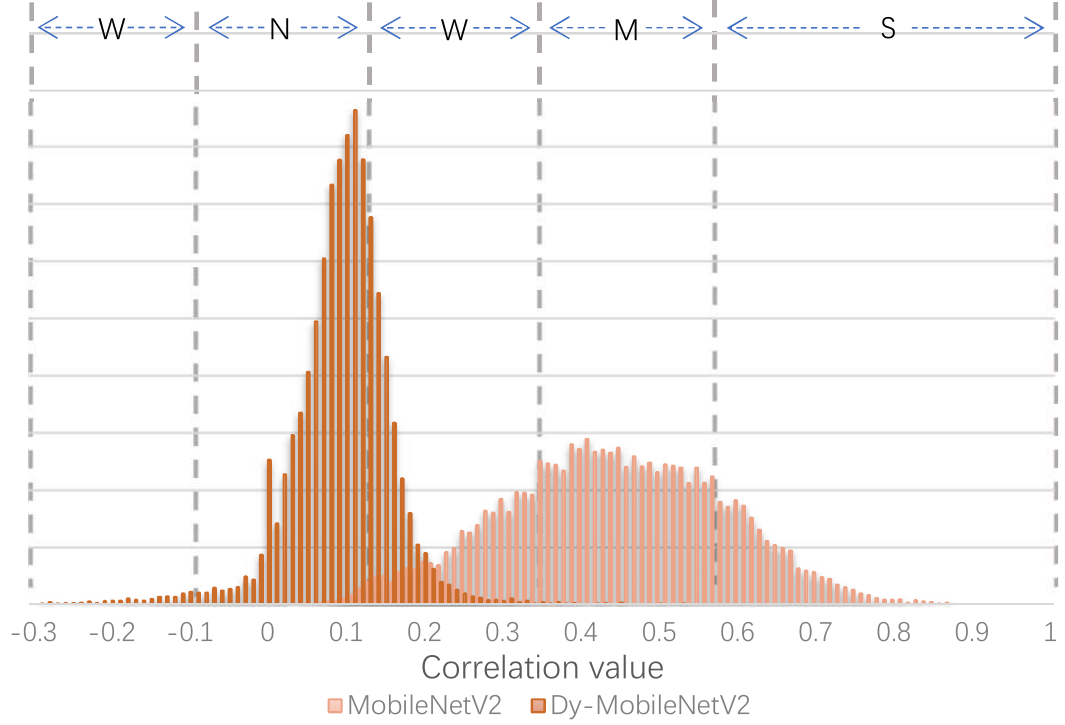}
\caption{Pearson product-moment correlation coefficient between feature maps, S, M, W, N denote strong, middle, weak and no correlation respectively. We can observe that compared with conventional kernels, the generated dynamic kernels have small correlation values.}
\label{dynamic_results}
\end{figure}
\begin{figure}[htb]
	\centering
	\includegraphics[width=0.8\linewidth]{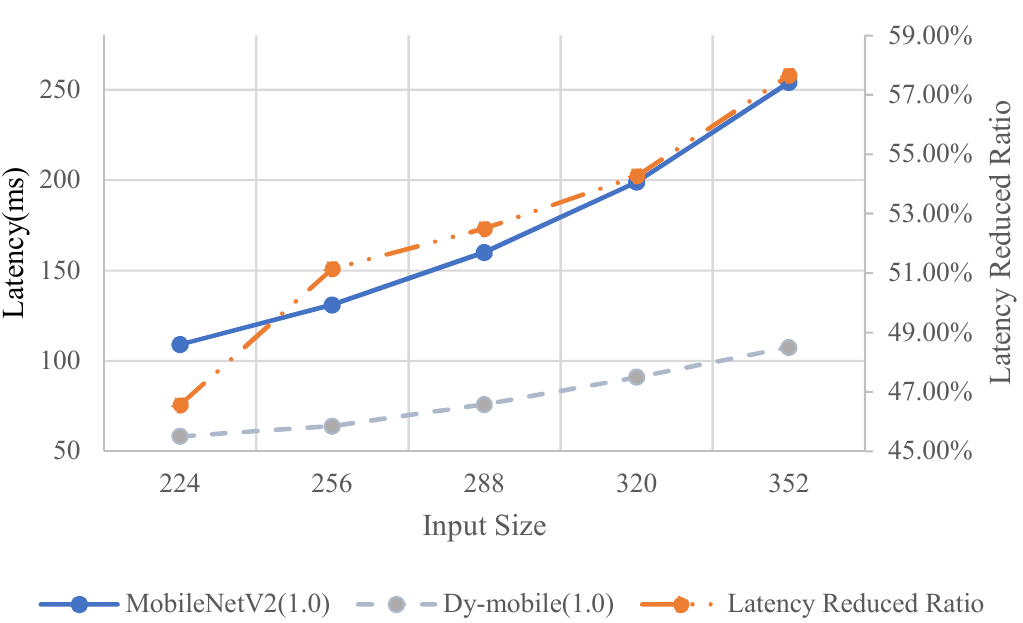}
	\caption{Latency for different input size.If we denote the latency of MobileNetV2(1.0),Dy-mobile as $L_{Fix}$ and $L_{Dym}$, then Latency Reduced Ratio is defined as $100\%-\frac{L_{Dym}}{L_{Fix}}$.}
	\label{latency_inputsize}
\end{figure}

\paragraph{Analysis of the dynamic kernel} Aside from the quantitative analysis, we also demonstrate the redundancy of the generated dynamic kernels compared with conventional kernels in Figure~\ref{dynamic_results}. We calculate the correlation between 160 feature maps output by the 7th stage for the original MobileNetV2(1.0) and Dy-MobileNetV2 (1.0) based on the validation set. Note that Dy-MobileNetV2 (1.0) is different with Dy-mobile(1.0). Dy-MobileNetV2(1.0) keeps the channels of each layer the same as the original one while replacing the conventional convolution with dynamic convolution. As shown in Figure~\ref{dynamic_results}, we can observe that the correlation distribution of dynamic kernels have more values distributed between $-0.1$ and $0.2$ compared with conventional convolution kernels, which indicates that the redundancy between dynamic convolution kernels are much smaller than the conventional convolution kernels.

\paragraph{Analysis of speed on the hardware}
We also analyze the inference speed of DyNet. We carry out experiments on the CPU platform (Intel(R) Core(TM) i7-7700 CPU @ 3.60GHz) with Caffe \citep{jia2014caffe}. We set the size of input as 224 and report the average inference time of 50 iterations. It is reasonable to set mini-batch size as 1, which is consistent with most inference scenarios. The results are shown in Table \ref{inference_table}. Moreover, the latency of fusing fixed kernels is independent with the input size, thus we expect to achieve a bigger acceleration ratio when the input size of networks becomes larger. We conduct experiments to verify this assumption, the results are shown in Figure \ref{latency_inputsize}. We can observe that the ratio of reduced latency achieved by DyNet gets bigger as the input size becomes larger. As shown in \citep{Tan2019EfficientNet}, a larger input size can make networks perform significantly better, thus DyNet is more effective in this scenario. 

We also analyze the training speed on the GPU platform. The model is trained with 32 NVIDIA Tesla V100 GPUs and the batch size is 2048. We report the average training time of one iteration in Table \ref{inference_table}. It is observed that the training speed of DyNet is slower, it is reasonable because we fuse feature maps rather than kernels according to Eq. \ref{eq9} in the training stage.

\setlength{\parskip}{0.1ex} 
\begin{table}[tbh]
\caption{Speed on the hardware.}
\label{inference_table}
\centering
\begin{tabular}{llll}
\toprule
Methods & Top-1 err. (\%) & Inference Time&Training Time\\
\midrule
MobileNetV2(1.0) & 28.00 &109.1ms&173ms\\
ResNet18  & 30.41 &90.7ms&170ms\\
ResNet50 & 23.67 &199.6ms&308ms\\
\midrule
Dy-mobile(1.0)  & 28.27 &58.3ms&250ms\\
Dy-ResNet18  & 31.01 &68.7ms&213ms\\
Dy-ResNet50 & 23.75 &135.1ms&510ms\\
\bottomrule
\end{tabular}
\end{table}

\subsection{Experiments on segmentation}
To verify the scalability of DyNet on other tasks, we conduct experiments on segmentation. Compared to the method Dilated FCN with ResNet50 as backbone \citep{Fu2018Dual}, Dilated FCN with Dy-ResNet50 reduces 69.3\% FLOPs while maintaining the MIoU on Cityscapes validation set. The result is shown in Table \ref{segmenttion}.

\begin{table}[tbh]
	\caption{Experiments of segmentation on Cityscapes val set.}
	\label{segmenttion}
	\centering
	\begin{tabular}{llll}
		\toprule
		Methods & BaseNet & GFLOPs & Mean IoU\%\\
		\midrule
		Dilated FCN\citep{Fu2018Dual} & ResNet50 & 310.8 &70.03\\
		Dilated FCN\citep{Fu2018Dual} & Dy-ResNet50 & 95.6 &70.48\\
		\bottomrule
	\end{tabular}
\end{table}

\subsection{Ablation study}
\paragraph{Comparison between convolution with conventional kernel and dynamic kernel}
We correspondingly design two baseline networks for Dy-mobile (1.0) and Dy-shuffle (1.5), denoted as Fix-mobile(1.0) and Fix-shuffle (1.5) respectively. Specifically, we remove the coefficient prediction module and dynamic generation module, using fixed convolution kernels directly, the channel number is kept changeless. The results are shown in Table \ref{baseline_table}, compare with baseline networks Fix-mobile (1.0) and Fix-shuffle (1.5), the proposed Dy-mobile (1.0) and Dy-shuffle (1.5) achieve absolute classification improvements by $5.19\%$ and $2.82\%$ respectively. This shows that directly decreasing the channel number to reduce computation cost influences the classification performance a lot. While the proposed dynamic kernel can retain the representation ability as much as possible.
\begin{table}[tbh]
	\caption{Ablation experiments results of convolution with conventional kernel and dynamic kernel.}
	\label{baseline_table}
	\centering
	\begin{tabular}{llll}
		\toprule
		Methods &MParams& MFLOPs & Top-1 err. (\%) \\
		\midrule
		Fix-mobile (1.0) &2.16& 129 & 33.57 \\
		Fix-shuffle (1.5) &2.47& 171 & 30.30 \\
		\midrule
		Dy-mobile (1.0) &7.36& 135 & 28.27 \\
		Dy-shuffle (1.5) &11.0& 180 & 27.48 \\
		\bottomrule
	\end{tabular}
\end{table}
\begin{table}[tbh]
	\caption{Ablation experiments on $g_t$.}
	\label{gt_table}
	\centering
	\begin{tabular}{llll}
		\toprule
		Methods &MParams& MFLOPs & Top-1 err. (\%) \\
		\midrule
		Fix-mobile(1.0) &2.16& 129 & 33.57 \\
		Dy-mobile(1.0, $g_t=2$) &3.58& 131 & 29.43 \\
		Dy-mobile(1.0, $g_t=4$) &5.47& 133 & 28.69 \\
		Dy-mobile(1.0, $g_t=6$) &7.36& 135 & 28.27 \\
		\bottomrule
	\end{tabular}
\end{table}
\begin{table}[tbh]
	\caption{Comparison for $g_t=1$ and $g_t=6$.}
	\label{gtse_table}
	\centering
	\begin{tabular}{llll}
		\toprule
		Methods &MParams& MFLOPs & Top-1 err. (\%) \\
		\midrule
		Dy-mobile (1.0, $g_t=1$) &2.64& 131 & 30.85 \\
		Dy-mobile (1.0, $g_t=6$) &7.36& 135 & 28.27 \\
		\midrule
		Dy-ResNet18 ($g_t=1$) &3.04& 553 & 33.8 \\
		Dy-ResNet18 ($g_t=6$) &16.6& 567 & 31.01 \\
		\bottomrule
	\end{tabular}
\end{table}

\paragraph{Effectiveness of $g_t$ for dynamic kernel}
The group size $g_t$ in Eq.~\ref{dy_conv} does not change the computation cost of DyNet but affects the performance of the network. Thus we provide an ablative study on $g_t$. We set $g_t$ as 2,4,6 for dy-mobile(1.0) respectively and the results are shown in Table~\ref{gt_table}. The performance of dy-mobile(1.0) becomes better when $g_t$ gets larger. It is reasonable because a larger $g_t$ means the number of kernels cooperated for obtaining one noise-irrelevant feature becomes larger.

When $g_t=1$, the coefficient prediction module can be regarded as merely learning the attention for different channels, which can improve the performance of networks as well~\citep{hu2018squeeze}. Therefore we provide ablative study for comparing $g_t=1$ and $g_t=6$ on Dy-mobile(1.0) and Dy-ResNet18. The results are shown in Table~\ref{gtse_table}. From the table we can see that, setting $g_t=1$ will reduce the Top-1 accuracy on ImageNet for Dy-mobile(1.0) and Dy-ResNet18 by 2.58\% and 2.79\% respectively. It proves that the improvement of our proposed dynamic networks does not only come from the attention mechanism.

\section{Conclusion}
\label{conclude}
In this paper, we propose a novel dynamic convolution method to adaptively generate convolution kernels based on image content, which reduces the redundant computation cost existed in conventional convolution kernels. Based on the proposed dynamic convolution, we design several dynamic convolution neural networks based on well-known architectures. The experiment results show that DyNet can reduce FLOPs remarkably while maintaining the performance or boost the performance while maintaining the computation cost. As future work, we want to further explore the redundancy phenomenon existed in convolution kernels, and find other ways to reduce computation cost, such as dynamically aggregate different kernels for different images other than fixed groups used in this paper.

\newpage
\bibliography{iclr2020_conference}
\bibliographystyle{iclr2020_conference}

\appendix
\newpage
\section{Appendix}\label{Appendix}

\subsection{Detailed analysis of our motivation}

We illustrate our motivation from a convolution with output $f(x)$, i.e.,
\begin{equation}
f(x)=x\otimes w,
\end{equation}
where $\otimes$ denotes the convolutional operator, $x\in R^n$ is a vectorized input and $w\in R^n$ means the filter. Specifically, the $i_{th}$ element of the convolution output $f(x)$ is calculated as:
\begin{equation}
f_i(x)=\langle  \,x_{(i)},w\rangle,
\end{equation}
where $\langle  \cdot,\cdot\rangle$ provides an inner product and $x_{(i)}$ is the circular shift of $x$ by $i$ elements. We define the index $i$ started from $0$.

We denote the noises in $x_{(i)}$ as $\sum_{j=0}^{d-1}{\alpha_j y_j}$, where $\alpha_j \in R$ and $\{y_0,y_1,...,y_{d-1}\}$ are the base vectors of noise space $\Psi$. Then the kernels in one convolutional layer can be represented as  $\{w_0,w_1,...,w_c\}$. The space expanded by $\{w_0,w_1,...,w_c\}$ is $\Omega$. We can prove if the kernels are trained until $\Psi \subset \Omega$,  then for each $w_k\notin \Psi$, we can get the noise-irrelevant $f_i(x^{white})=\langle x^{white}_{(i)},w_k\rangle$ by the cooperation of other kernels ${w_0,w_1,...}$.

Firstly $x_{(i)}$ can be decomposed as:
\begin{equation}
x_{(i)}=\bar{x}_{(i)}+\beta w_k+\sum_{j=0}^{d-1}{\alpha_j y_j},
\end{equation}
where $\beta \in R$ and $\bar{x} \in R^n$ is vertical to $w_k$ and ${y_j}$.

For concision we assume the norm of $w_k$ and $y_j$ is 1. Then,
\begin{equation}
\begin{aligned}
f_i(x)=\langle   x_{(i)},w_k\rangle=\langle  \bar{x}_{(i)}+\beta w_k+\sum_{j=0}^{d-1}{\alpha_j y_j},w_k\rangle=
\beta\langle   w_k,w_k\rangle +\sum_{j=0}^{d-1}{\alpha_j \langle   y_j,w_k\rangle}
\label{eq4}
\end{aligned}
\end{equation}

When there is no noise, i.e. $\alpha_j=0$ for $j=0,1,...,d-1$, the white output $f_i(x^{white})$ becomes:
\begin{equation}
f_i(x^{white})=\langle   x^{white}_{(i)},w_k\rangle=\langle  \bar{x}_{(i)}+\beta w_k,w_k\rangle=\beta \langle   w_k,w_k\rangle=\beta.
\end{equation}

It is proved in the Appendix~\ref{appendixa2} that:
\begin{equation}
\begin{aligned}
f_i(x^{white})=\langle   a_{00}w_k+\sum_{t}{\beta_t w_t},x_{(i)}\rangle=(a_{00}+\beta_k)\langle   w_k,x_{(i)}\rangle+\sum_{t \ne k}{\beta_t \langle   w_t,x_{(i)}}\rangle,\label{eq6}
\end{aligned}
\end{equation}
where $\beta_0,...,\beta_c$ is determined by the input image.

Eq. \ref{eq6} is fulfilled by linearly combine convolution output $\langle   w_k,x_{(i)}\rangle$ and $\langle   w_t,x_{(i)}\rangle$ for those $\beta_t \ne 0$ in the following layers. Thus if there are $N$ coefficients in Eq.~\ref{eq6} that are not 0, then we need to carry out $N$ times convolution operation to get the noise-irrelevant output of kernel $w_t$, this causes redundant calculation.

In Eq. \ref{eq6}, we can observe that the computation cost can be reduced to one convolution operation by linearly fusing those kernels to a dynamic one:
\begin{equation}
\begin{aligned}
\widetilde{w}=(a_{00}+\beta_k)w_k+\sum_{t \ne k,\beta_t \ne 0}{\beta_t w_t}\\
f_i(x^{white})=\langle  \widetilde{w},x_{(i)}\rangle.\label{eq7}
\end{aligned}
\end{equation}

In Eq. \ref{eq7}, the coefficients $\beta_0,\beta_1,...$ is determined by $\alpha_0,\alpha_1,...$, thus they should be generated based on the input of network. \textit{This is the motivation of our proposed dynamic convolution.}

\subsection{Proving of Eq.~\ref{eq6}}
\label{appendixa2}
We denote $g_{ij}(x)$ as $\langle x_{(i)},y_j\rangle $, $j=0,1,...,d-1$. Then,
\begin{equation}
g_{ij}(x)=\langle x_{(i)},y_j\rangle =\langle  \bar{x}_{(i)}+\beta w_k+\sum_{t=0}^{d-1}{\alpha_t y_t},y_j\rangle =\beta\langle w_k,y_j\rangle  +\sum_{t=0}^{d-1}{\alpha_t \langle y_t,yj\rangle }.
\label{eq11}
\end{equation}

By summarize Eq.~\ref{eq4} and Eq.~\ref{eq11}, we get the following equation:

\begin{equation}
\setlength{\arraycolsep}{0.4pt}
\left[
	\begin{matrix}
	\langle w_k,w_k\rangle & \langle y_0,w_k\rangle & \langle y_1,w_k\rangle &...&\langle y_{d-1},w_k\rangle & \\
	\langle w_k,y_0\rangle & \langle y_0,y_0\rangle & \langle y_1,y_0\rangle &...&\langle y_{d-1},y_0\rangle & \\
	\langle w_k,y_1\rangle & \langle y_0,y_1\rangle & \langle y_1,y_1\rangle &...&\langle y_{d-1},y_1\rangle & \\
	\vdots & \vdots & \vdots & ... & \vdots & \\
	\langle w_k,y_{d-1}\rangle & \langle y_0,y_{d-1}\rangle  & \dots & ... & \langle y_{d-1},y_{d-1}\rangle & \\
	\end{matrix}
\right]
\left[
	\begin{matrix}
	\beta \\
	\alpha_0 \\
	\alpha_1 \\
	\vdots \\
	\alpha_{d-1} \\
	\end{matrix}
\right]=
\left[
	\begin{matrix}
	f_i(x) \\
	g_{i0}(x) \\
	g_{i1}(x) \\
	\vdots \\
	g_{i{(d-1)}}(x) \\
        \end{matrix}
\right],
\end{equation}

We simplify this equation as:

\begin{equation}
A\vec{x}=\vec{b}.
\end{equation}

Because $w_k \notin \Psi$, we can denote $w_k$ as:
\begin{equation}
w_k=\gamma_{\perp}w_\perp+\sum_{j=0}^{d-1}{\gamma_{j}y_j},
\end{equation}
where $w_\perp$ is vertical to $y_0,...,y_{d-1}$ and $\gamma_{\perp}\ne0$.

moreover because $|w_k|=1$ ,thus
\begin{equation}
|\gamma_\perp|^2+\sum_{j=0}^{d-1}{|\gamma_j|^2}=1.
\end{equation}
It can be easily proved that:
\begin{equation}
\setlength{\arraycolsep}{0.4pt}
A=\left[
	\begin{matrix}
	1 & \gamma_0 & \gamma_1 & ... & \gamma_{d-1}& \\
	\gamma_0 & 1 & 0 & ... & 0 & \\
	\gamma_1 & 0 & 1 & ... & 0 & \\
	\vdots & \vdots & \vdots & ... & \vdots & \\
	\gamma_{d-1} & 0 & \dots & ... & 1 & \\
	\end{matrix}
\right].
\end{equation}

thus,
 \begin{equation}
 \begin{aligned}
\setlength{\arraycolsep}{0.4pt}
|A|=&
\left|
	\begin{matrix}
	1 & \gamma_0 & \gamma_1 & ... & \gamma_{d-1} & \\
	\gamma_0 & 1 & 0 & ... & 0 & \\
	\gamma_1 & 0 & 1 & ... & 0 & \\
	\vdots & \vdots & \vdots & ... & \vdots & \\
	\gamma_{d-1} & 0 & \dots & ... & 1 & \\
	\end{matrix}
\right|\\
=&\left|
	\begin{matrix}
	1-\gamma_0^2 & 0 & \gamma_1 & ... & \gamma_{d-1} & \\
	\gamma_0 & 1 & 0 & ... & 0 & \\
	\gamma_1 & 0 & 1 & ... & 0 & \\
	\vdots & \vdots & \vdots & ... & \vdots & \\
	\gamma_{d-1} & 0 & \dots & ... & 1 & \\
	\end{matrix}
\right|\\
=&\left|
	\begin{matrix}
	1-\gamma_0^2-\gamma_1^2 & 0 & 0 & ... & \gamma_{d-1} & \\
	\gamma_0 & 1 & 0 & ... & 0 & \\
	\gamma_1 & 0 & 1 & ... & 0 & \\
	\vdots & \vdots & \vdots & ... & \vdots & \\
	\gamma_{d-1} & 0 & \dots & ... & 1 & \\
	\end{matrix}
\right|\\
=&\left|
	\begin{matrix}
	1-\gamma_0^2-\gamma_1^2-...-\gamma_{d-1}^2 & 0 & 0 & ... & 0 & \\
	\gamma_0 & 1 & 0 & ... & 0 & \\
	\gamma_1 & 0 & 1 & ... & 0 & \\
	\vdots & \vdots & \vdots & ... & \vdots & \\
	\gamma_{d-1} & 0 & \dots & ... & 1 & \\
	\end{matrix}
\right|
\\=&\left|
	\begin{matrix}
	\gamma_{\perp}^2 & 0 & 0 & ... & 0 & \\
	\gamma_0 & 1 & 0 & ... & 0 & \\
	\gamma_1 & 0 & 1 & ... & 0 & \\
	\vdots & \vdots & \vdots & ... & \vdots & \\
	\gamma_{d-1} & 0 & \dots &... & 1 & \\
	\end{matrix}
\right|\\
=&\gamma_{\perp}^2\ne0.
\end{aligned}
\end{equation}

thus,
\begin{equation}
\vec{x}=A^{-1}\vec{b}.
\end{equation}
If we denote the elements of the first row of $A^{-1}$ as $a_{00},a_{01},...,a_{0d}$, then
\begin{equation}
\begin{aligned}
f_i(x^{white})=\beta&=a_{00}f_i(x)+\sum_{j=0}^{d-1}{a_{0(j+1)}}g_{i,j}(x)\\&=a_{00}\langle w_k,x_{(i)}\rangle +\sum_{j=0}^{d-1}{a_{0(j+1)}}\langle y_j,x_{(i)}\rangle \\ &
=\langle a_{00}w_k+\sum_{j=0}^{d-1}{a_{0(j+1)}}y_j,x_{(i)}\rangle .
\end{aligned}
\end{equation}

Because $\Psi \subset \Omega$, there exists ${\{\beta_t\in R |t=0,1,...,c}\}$ that
\begin{equation}
\sum_{j=0}^{d-1}{a_{0(j+1)}}y_j=\sum_{t}{\beta_t w_t}.
\end{equation}

Then,
\begin{equation}
\begin{aligned}
f_i(x^{white})=\langle a_{00}w_k+\sum_{t}{\beta_t w_t},x_{(i)}\rangle =(a_{00}+\beta_k)\langle w_k,x_{(i)}\rangle +\sum_{t \ne k}{\beta_t\langle w_t,x_{(i)}}\rangle ,
\end{aligned}
\end{equation}
\end{document}